\documentclass[letterpaper, 10 pt, conference]{ieeeconf}  

\IEEEoverridecommandlockouts                              

\usepackage{graphicx}
\usepackage{amsmath}
\usepackage{}
\usepackage{cite}
\usepackage{subfiles}
\usepackage{booktabs}
\usepackage{algorithm}
\usepackage[table]{xcolor}
\usepackage{algorithmic}
\usepackage{xcolor} 
\usepackage{tcolorbox}
\usepackage{hyperref}
\usepackage{cleveref}
\tcbuselibrary{skins}

\newcounter{example}

\title{\LARGE \bf
MCCoder: Streamlining Motion Control with LLM-Assisted Code Generation and Rigorous Verification
}

\author{Yin Li$^{1}$, Liangwei Wang$^{1}$, Shiyuan Piao$^{1}$, Boo-Ho Yang$^{2}$, Ziyue Li$^{3}$, Wei Zeng$^{1}$, Fugee Tsung$^{1,4}$
\thanks{$^{1}$YL, LW, SP, WZ and FT are with the Thrust of Data Science and Analytics, The Hong Kong University of Science and Technology (Guangzhou), Guangzhou, China.
        {\tt\small \{yligt, lwang344, spiao277\}@connect.hkust-gz.edu.cn, \{weizeng, season\}@hkust-gz.edu.cn}}%
\thanks{$^{4}$FT is also with the Department of Industrial Engineering and Decision Analytics, The Hong Kong University of Science and Technology, Hong Kong SAR, China.}%
\thanks{$^{2}$BY is with MOVENSYS Inc., Seongnam-si, Republic of Korea.}%
\thanks{$^{3}$ZL is with University of Cologne, Cologne, Germany.}%
}

\usepackage{algorithm}
\usepackage{algorithmic}
\usepackage{booktabs}
\usepackage{listings}
\usepackage{multirow}
\begin{document}

\maketitle
\thispagestyle{empty}
\pagestyle{empty}


\begin{abstract}

Large Language Models (LLMs) have demonstrated significant potential in code generation. However, in the factory automation sector—particularly motion control—manual programming, alongside inefficient and unsafe debugging practices, remains prevalent. This stems from the complex interplay of mechanical and electrical systems and stringent safety requirements. Moreover, most current AI-assisted motion control programming efforts focus on PLCs, with little attention given to high-level languages and function libraries.
To address these challenges, we introduce MCCoder, an LLM-powered system tailored for generating motion control code, integrated with a soft-motion controller. MCCoder improves code generation through a structured workflow that combines multitask decomposition, hybrid retrieval-augmented generation (RAG), and iterative self-correction, utilizing a well-established motion library. Additionally, it integrates a 3D simulator for intuitive motion validation and logs of full motion trajectories for data verification, significantly enhancing accuracy and safety.
In the absence of benchmark datasets and metrics tailored for evaluating motion control code generation, we propose MCEVAL, a dataset spanning motion tasks of varying complexity. Experiments show that MCCoder outperforms baseline models using Advanced RAG, achieving an overall performance gain of 33.09\% and a 131.77\% improvement on complex tasks in the MCEVAL dataset.
MCCoder is publicly available at \href{https://github.com/MCCodeAI/MCCoder}{https://github.com/MCCodeAI/MCCoder}.

\end{abstract}



\section{INTRODUCTION}

\begin{figure}[t]
    \centering
    \includegraphics[width=1\columnwidth]{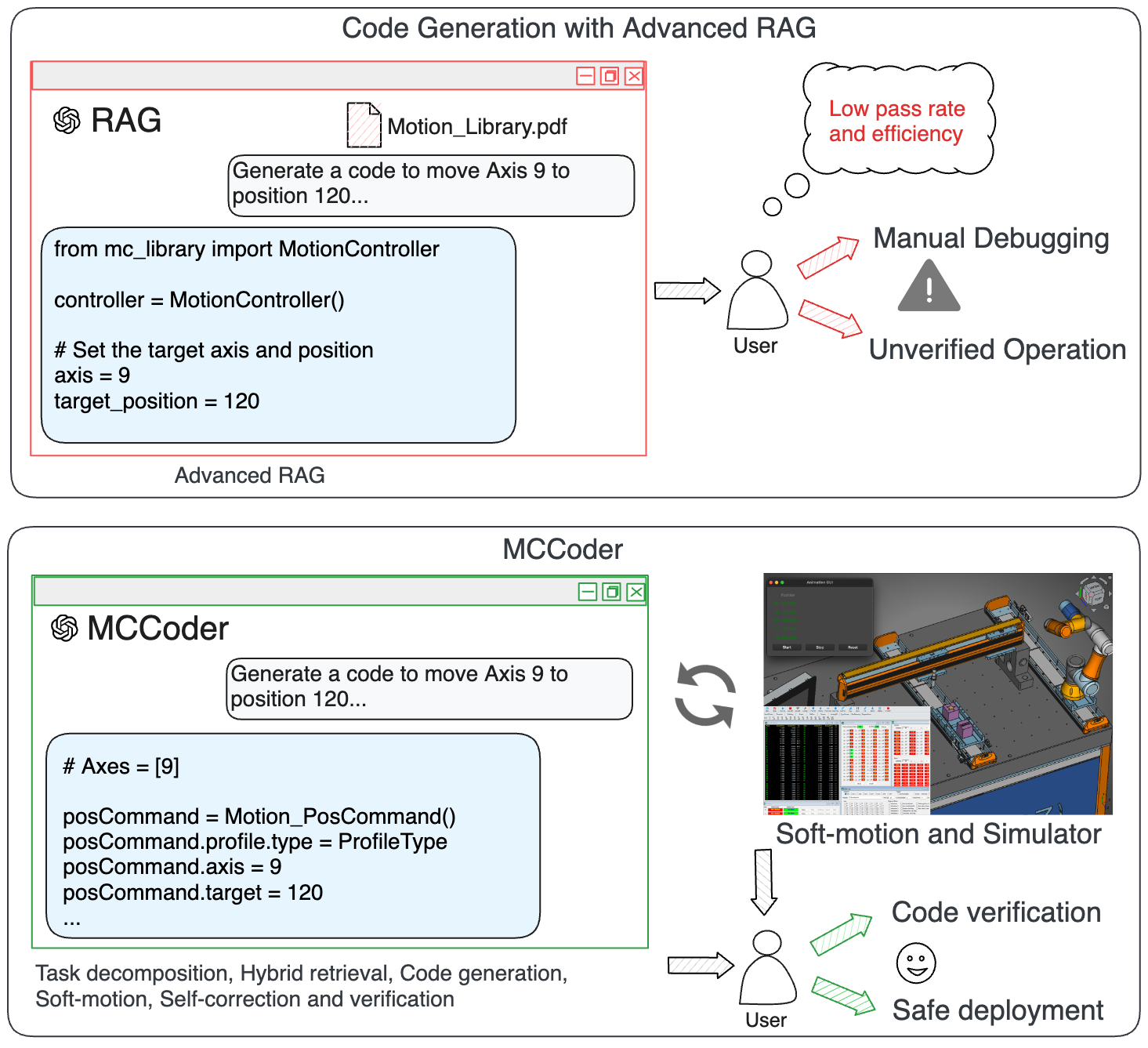}
    \caption{Comparison between Advanced RAG
and LLM-assisted MCCoder. Code generation with Advanced RAG has a low pass rate and efficiency, requiring extensive debugging and posing safety concerns. In contrast, MCCoder generates code through task decomposition, self-correction, and soft-motion with simulator, ensuring data verification and safe machine deployment.}
    \vspace{-20pt}
    \label{fig:aragvsmccoder}
\end{figure}

Motion control, a fundamental element of factory automation, has greatly improved industrial processes, evolving from the introduction of numerical control (NC) machines in the 1950s to the advanced AI-powered robots and semiconductor equipment of today.While motion control spans a range of programming approaches (e.g., CAD/CAM, PLC, and robot teaching), this paper concentrates specifically on the domain of motion API invocation. Presently, software engineers in this field primarily depend on manual programming and debugging of motion control systems, driven by the intricate nature of manufacturing processes and critical machinery safety considerations. Certain motion libraries include over 1,000 APIs with override arguments and more than 100 synchronous motion channels, creating a steep learning curve and limiting the practicality of automated programming tools. In addition, engineers must iteratively debug in coordination with electrical and mechanical components to enhance motion performance and address potential hazards to both human operators and machinery during operations.

The recent spotlight on LLM-based code generation has unveiled impressive ability to address programming challenges \cite{Du2023ClassEval:}. Although traditional code generation in automation control often relied on model-driven or rule-based approaches, recent advancements have begun to explore LLM-based methods. For example, a retrieval-augmented approach \cite{koziolek2024llm} was proposed to generate IEC 61131-3 programs, while LLM4PLC \cite{fakih2024llm4plc} used user feedback and external verification tools to guide the generation of LLM-based PLC codes. 

However, current attempts to use LLMs for motion control primarily focus on standardized PLC languages, neglecting higher-level languages like Python or C++ commonly used in complex automation equipment. Furthermore, there is a lack of thorough research on verifying the safety and effectiveness of generated control code, as well as a shortage of suitable datasets for evaluation.

To address these limitations, we propose MCCoder, which utilizes LLMs to generate Python code for motion control based on natural language instructions. This system employs a motion library to decompose complex tasks, generate code, and use soft-motion for simulation running and data logging for verification. In addition, we build the MCEVAL dataset to evaluate the generated code. \Cref{fig:aragvsmccoder} highlights a comparison between code generation with Advanced RAG and LLM-assisted MCCoder. Experiments demonstrated that MCCoder improves code generation performance, achieving an overall improvement of 33.09\% and a significant increase of 131.77\% in complex tasks in the MCEVAL dataset.

The primary contributions of our work are as follows.

\begin{itemize}
    \item \textbf{MCCoder System:} MCCoder is an AI-assisted code generation and verification system specifically designed for motion control in factory automation. It integrates task decomposition, hybrid retrieval, code generation, soft-motion, self-correction, and data verification to ensure code safety and effectiveness in handling complex real-world programming challenges. Soft-motion with LLM plays a central role, transforming the traditional approach of manual programming or LLM-based coding, and human debugging into an automated workflow.
    \item \textbf{Evaluation Dataset and Metrics:} Existing code generation datasets and metrics are not designed for the motion control domain. We introduce MCEVAL, the first benchmark dataset and evaluation metrics specifically tailored for LLM-assisted motion control code. MCEVAL comprises 186 motion control programming tasks, ranging from simple to complex, covering most common functionalities and domains in motion control. It carefully considers function diversity, potential ambiguities, and interactions among multiple subtasks, while its evaluation metrics account for both motion endpoints and trajectory assessment. MCEVAL fills the gap in general-purpose code generation benchmarks and the motion control field, serving as a valuable reference for professionals.
\end{itemize}

\section{PRELIMINARY}

MCCoder utilizes a soft-motion architecture to generate Python code for motion control by calling APIs from WMX3 motion library.

\subsection{Soft-motion}

A motion controller is typically used to control the motion of servo motors, axes, and I/Os in machines, relying on dedicated hardware architecture. In contrast, a soft-motion system features a flexible and software-based architecture that runs on a general-purpose PC without requiring specialized hardware. It leverages a CPU core with a real-time OS and complex motion algorithms, ensuring high performance and scalability.

MCCoder interacts with the soft-motion via a DLL. The soft-motion system first executes the control code in a simulation engine for data logging and verification. Once validated, the real-time engine transmits control commands to the machine through cyclic fieldbus communication. Moreover, its PC-based software architecture makes soft-motion particularly well suited for integrating advanced functionalities, such as LLMs and other AI technologies.

\subsection{Control Code and Motion Library}

Unlike standardized PLC programming, such as IEC 61131-3, control code programming in Python offers greater flexibility but is harder to constrain within a fixed template. In this research, we utilize the well-known WMX3 motion library within a soft-motion controller, which provides over 1,000 APIs capable of managing up to 128 axes and 256 independent task channels. These APIs cover a wide range of motion control functionalities, including fieldbus communication, digital and analog I/O control, and diverse trajectory, position, velocity, and current control for servo axes. Notably, these API functions exhibit a complex structure: the same function name may correspond to multiple overrides and parameter classes, requiring even experts with extensive industry experience to carefully distinguish their differences, combinations, and application scenarios—a significant challenge for AI-driven programming. \Cref{exam1} presents a sample code demonstrating how to move axes to specified positions, encompassing the INITIALIZE, MOTION, and CLOSE procedures.

\begin{tcolorbox}[
    colback=white,
    colframe=gray,
    title={Example 1. \tt Motion Control Sample Code}
    \setlength{\leftskip}{-1em}]
    \refstepcounter{example}\label{exam1}
\begin{algorithmic}[0]
    \setlength{\leftskip}{-1.5em} 
    \fontsize{10pt}{10pt}\selectfont
    \STATE $Axes$ = [$A_i$] 
    \STATE $IO_{Ins}$ = [$I_i$] 
    \STATE $IO_{Outs}$ = [$O_i$]  
    
    \STATE \textbf{INITIALIZE():}
    \STATE $MCLib.CreateDevice()$
    \STATE $MCLib.StartCommunication()$
    \STATE $MCLib.SetandStartLog(Axes, IO_{Ins}, IO_{Outs})$

    \STATE \textbf{MOTION():}
    \FOR{$i$ in $Axes$}
    \STATE $posCommand = MotionPosCommand()$
    \STATE $posCommand.profile.type = ProfileType_i$
    \STATE $posCommand.axis = i$
    \STATE $posCommand.target = Pos_i$
    \STATE $MCLib.motion.StartPos(posCommand)$
    \STATE $MCLib.motion.Wait(i)$
    \ENDFOR
    \STATE \textbf{CLOSE():}
    \STATE $MCLib.CloseDevice()$
    \STATE \textbf{RETURN:} Success or error code
\end{algorithmic}
\end{tcolorbox}

\section{RELATED WORK}

\subsection{Code Generation with LLMs and Strategies}

Recent advancements in LLMs, such as OpenAI's Codex, GitHub's Copilot, Google's Gemini, and Meta's CodeLlama, have significantly enhanced code generation, improving both productivity and code quality \cite{wang2023codet5+}. Trained on diverse codebases with advanced NLP techniques, these models excel in code completion, refinement, and debugging \cite{Li2023Enabling}. Benchmarks like HumanEval, MBPP, and CodeXGLUE reveal that these models are nearing or exceeding human-level performance in coding tasks \cite{Zhuo2023Large}. Enhanced model architectures with larger context windows and sizes allow them to tackle more complex tasks \cite{Jiang2023Self-planning}. Code generation strategies in LLMs include task planning \cite{Ruan2023TPTU:}, in-context learning \cite{brown2020languagemodelsfewshotlearners}, Chain-of-Thought prompting \cite{wei2023chainofthoughtpromptingelicitsreasoning}, RAG, and post-process methods like self-consistency, DIN-SQL \cite{Pourreza2023DIN-SQL:}, and DAIL-SQL \cite{Chen2023Teaching}, all of which improve the efficiency, accuracy, and reliability of these models.

\subsection{Code Generation in Motion Control}

Experts in industrial automation have explored methods to automate programming tasks, including model-driven development environments \cite{Hästbacka2011Model-driven} and rule-based systems like automated Matlab-to-C++ translators \cite{yang2022m2coder}. 13 code generation methods for control logic has been classified since 2004 \cite{Koziolek2020A}, but these traditional methods were limited to small-scale applications. Large Language Models (LLMs) offer significant advantages for large-scale applications. Researchers are now using LLMs for innovative code generation methods. For example, a retrieval-augmented approach was proposed for generating IEC 61131-3 Structured Text programs using GPT-4, LangChain, and OpenPLC, validated through expert simulations \cite{koziolek2024llm}. Similarly, LLM4PLC was developed as a user-guided iterative pipeline using syntax checkers and LLM fine-tuning for PLC programming \cite{fakih2024llm4plc}. These methods represent a promising start for automating control code generation.


\section{METHODOLOGY}

\begin{figure}[t]
    \centering
    \includegraphics[width=1\linewidth]{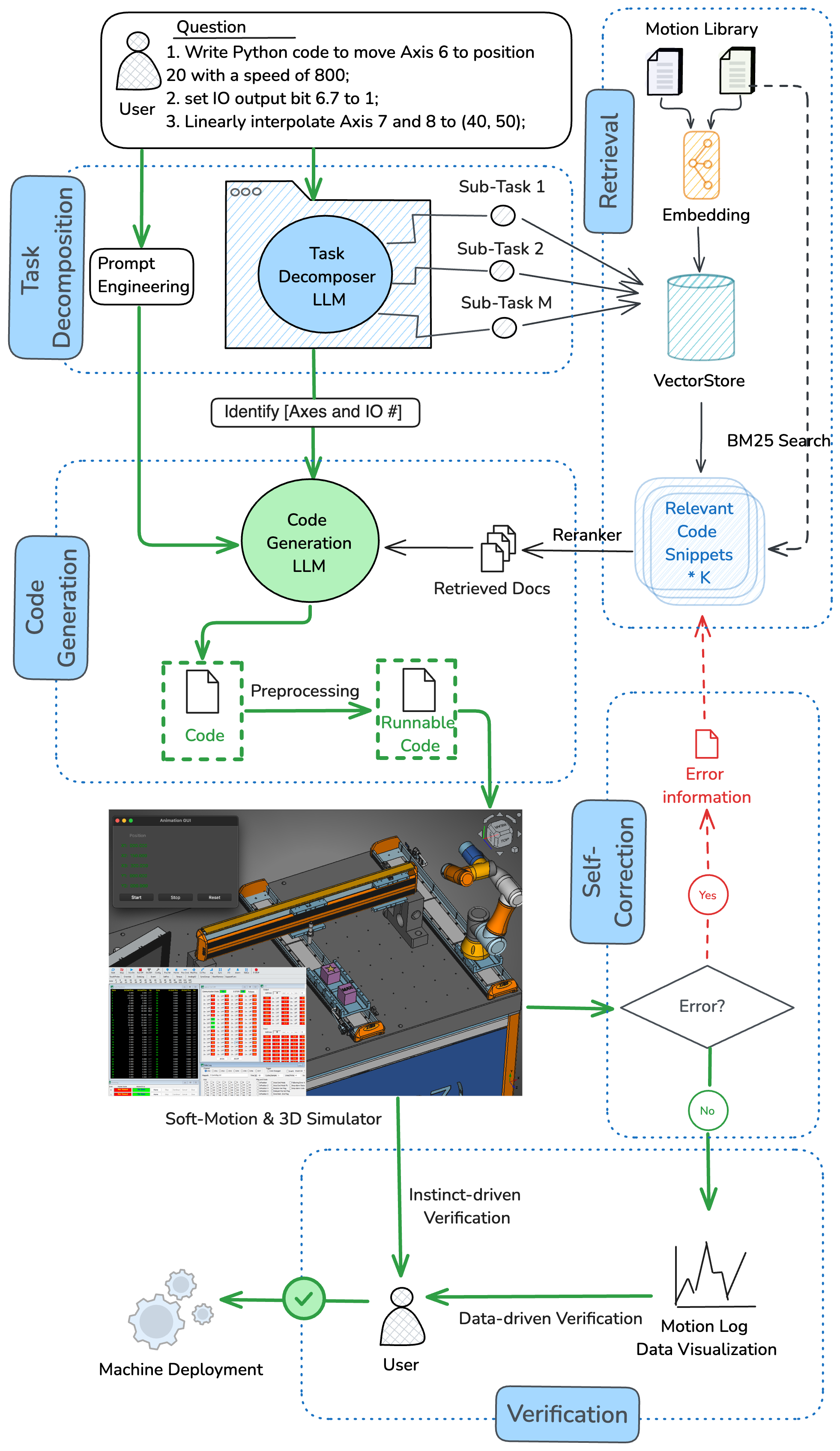}
    \vspace{-20pt}
    \caption{Overview of the MCCoder system. It integrates task decomposition, hybrid retrieval-augmented code generation, simulation, and self-correction with soft-motion system to enhance control code programming. It employs instinct-driven and data-driven verification before machine execution and deployment, greatly improving the efficiency and safety of control code generation.} 
    \label{fig:mccoder}
    \vspace{-20pt}
\end{figure}

We developed the MCCoder system to tackle the code generation challenges in motion control. Figure \ref{fig:mccoder} shows an overview of the MCCoder system. It involves six modules: task decomposition, hybrid retrieval, control code generation, soft-motion, self-correction and data verification.

\subsection{Task Decomposition}
The Task Decomposition Module, powered by an LLM, accepts user questions phrased as motion control tasks in natural language. These questions may involve multiple interconnected control tasks with complex dependencies and mutual triggering relationships. To enhance the LLM's comprehension and facilitate the retrieval of pertinent example code and documentation, this module breaks down the question into discrete subtasks for subsequent processing. Additionally, it extracts axis numbers and input/output identifiers from the question, preparing the necessary preprocessing for the downstream soft-motion system.

\subsection{Hybrid Retrieval}
The retrieval module aims to efficiently locate relevant information from extensive documentation and sample code. Thousands of pages of explanatory documents and code examples are chunked separately, using distinct strategies to preserve code integrity. Prior studies have introduced various RAG techniques, such as native RAG, advanced RAG, and modular RAG \cite{gao2023retrieval}. In our case, accurately retrieving correct API overrides and example code is challenging due to high similarity among motion APIs. Thus, we employ Sparse Retrieval (BM25), Dense Retrieval (embedding-based VectorStores), and a re-ranker to select the top-$k$ (set as 6) results, effectively matching user queries with the most relevant content.

\subsection{Control Code Generation}
This process employs an LLM combined with retrieved motion-specific sample code and documentation to generate control code. Prompt engineering is essential for enforcing proper code formatting, standards, and critical constraints.
The generated code undergoes preprocessing, including initialization steps such as enabling the corresponding servo axes, configuring IO, and setting up logging prior to execution. Termination logging and process closure instructions are appended after the code. This ensures the code is properly formatted and runnable before deployment to the soft-motion system.

\subsection{Soft-Motion}
The soft-motion system comprises a real-time engine for controlling physical machines and a simulation engine. MCCoder transmits the finalized code to the soft-motion system, which executes it in the simulation engine and provides feedback, including correct results or error messages for self-correction.
Additionally, the simulation engine logs motion data from axes and IO at 1ms intervals throughout execution, facilitating process verification and endpoint validation.
  
\subsection{Self-Correction}
If the soft-motion system detects a syntax or API error, MCCoder retrieves relevant documentation, regenerates the code, and re-executes it through an iterative self-correction process. Once execution is error-free, the soft-motion system logs the full trajectory data for subsequent verification. This synergy establishes a closed-loop feedback mechanism, enabling iterative code refinement and secure validation.

\subsection{Data Verification}
The 3D model of a machine can be imported into a 3D simulator, enabling instinct-driven verification by visualizing real-time motion during the execution of motion code in the soft-motion system and alerting users to potential collisions. Additionally, the soft-motion engine performs data-driven verification by logging comprehensive execution data every 1ms, capturing all axis and IO details to facilitate fine-grained analysis and ensure process accuracy.

\section{EVALUATION DATASET}
Existing code generation datasets, like HumanEval \cite{chen2021evaluating} and MBPP EvalPlus \cite{evalplus}, focus on general-purpose code evaluated via unit tests. The generation of control code requires execution in motion controllers to record the endpoint and trajectory data. MCEVAL evaluation dataset fills this gap for the automation industry. We meticulously constructed MCEVAL, verifying each task both manually and through soft-motion simulations to ensure that the programming tasks are well-formed.

\subsection{Construction}

MCEVAL consists of 186 tasks selected to offer a diverse range of motion control programming challenges. It encompasses the most commonly used motion control functions in the library, spanning from simple to complex. These functions cover point-to-point motion, linear, circular, and helical interpolation, splines, short-segment look-ahead, compensation, and event-driven interactions between axes and IOs across various profiles, along with their combinations and dependencies. The tasks also account for inaccuracies in user input. Each task is defined by a TaskId, Instruction, CanonicalCode, and Difficulty, with their properties detailed in \Cref{tab:MCEVAL}.

To replicate a human engineer's problem-solving skills, we categorized the tasks into three difficulty levels, with examples provided in \Cref{exam3}. Difficulty Level 1 involves calling standard APIs with straightforward arguments as outlined in the documentation. Difficulty Level 2 features APIs with more complex parameter types, including easily confused overrides, as well as potential user errors such as typos or improper parameter instructions. LLMs may err here but should leverage error feedback or retrieved content for self-correction. Difficulty Level 3 requires integrating multiple motion tasks, not merely stacking simple ones, but combining them into a cohesive set linked by events and dependencies. Solving these demands robust retrieval of all subtasks and a comprehensive understanding of the overall task. These levels mirror real-world motion control scenarios, progressing from simple to complex.

\begin{table}[t]
    \caption{Property and Structure of MCEVAL Dataset}
    \label{tab:MCEVAL}
    \centering
    \resizebox{\columnwidth}{!}{
        \begin{tabular}{cccc}
            \toprule
            Tasks    & Numbers & Instruction Length & Cannonical Code Length \\
            \midrule
            Difficulty Level 1    & 84 & 267 & 2268   \\
            Difficulty Level 2    & 56 & 289 & 1351   \\
            Difficulty Level 3    & 46 & 406 & 2926   \\         
            \midrule
            Total/\underline{Mean}         & 186  & \underline{273} & \underline{2145}   \\
            \bottomrule
        \end{tabular}
    }
\vspace{-20pt}
\end{table}






\begin{tcolorbox}[
    colback=white,         
    colframe=gray,        
    title={Example 2. {\tt  MCEVAL Dataset}}
    \setlength{\leftskip}{-1em}]
    \refstepcounter{example}\label{exam3} 
\begin{algorithmic}[0]
    \setlength{\leftskip}{-1.5em} 
    \fontsize{10pt}{10pt}\selectfont
    
\STATE // \textbf{Difficulty: 1}
\STATE \textbf{Instruction:} Write Python code to move axis 1 to position 130.2 at a speed of 1060, and acceleration of 11000.

\vspace{1ex} 
\hrule 
\vspace{1ex} 

\STATE // \textbf{Difficulty: 2}
\STATE \textbf{Instruction:} Write Python code to move Axis 9 to the position 90 at a speed of 1000, jerkAccRatio of 0.5, end Velocity of 0, using a Jerk-Ratio profile.

\vspace{1ex} 
\hrule 
\vspace{1ex} 

\STATE // \textbf{Difficulty: 3}
\STATE \textbf{Instruction:} Write Python code to set the input event to monitor if the DistanceToTarget of Axis 3's movement is 500, then move Axis 1 to the position -200 at a speed of 1000. Move Axis 3 to 1200...

\end{algorithmic}
\end{tcolorbox}

\subsection{Metrics}

To comprehensively assess the generation of code on the evaluation dataset, the following metrics are chosen based on control expertise.

\subsubsection{First Time Pass Rate (FTPR)}

FTPR is calculated as \Cref{eq:ftpr} representing the proportion of codes that pass the test on the first attempt \cite{loftin2020closer}. Pass@k is commonly used in code generation metrics and repeats many times to eliminate bias, but practical users prefer code that works correctly on the first try. We chose this straightforward measure to align with real-world use cases for control code.

\begin{equation}
FTPR =  \frac{N_{Passed}}{N_{Total}} * 100
\label{eq:ftpr}
\end{equation}

\subsubsection{MatchEndPoints and DTW}
The soft-motion engine generates log data every 1ms during code execution. Log files from the canonical code in MCEVAL dataset and the generated code are compared. MatchEndPoints compares only the end points, while Dynamic Time Warping (DTW) compares all trajectory points, measuring the similarity of time series data. In motion control, it depends on specific scenario whether to check end points or trajectories, for instance, interpolation motions should match trajectories, while point-to-point motions only need endpoint matching. We will measure both FTPR (MatchEndPoints) and FTPR (DTW).


\section{EXPERIMENTS}
\subsection{Experimental Settings}
\begin{figure}[h]
    \centering
    \includegraphics[width=0.9\linewidth]{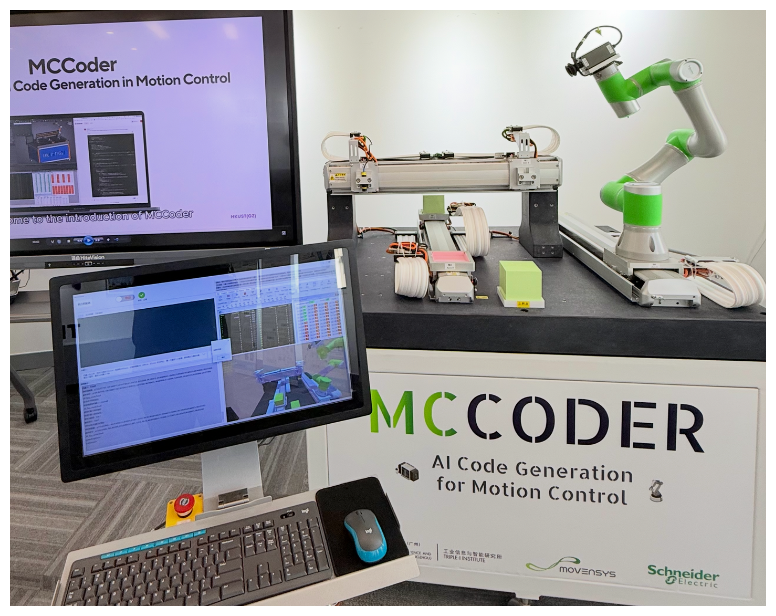}
    \vspace{-10pt}
    \caption{MCCoder Demo Machine.} 
    \label{fig:mccoderdemomachine}
    \vspace{-15pt}
\end{figure}

We established a demo machine, as depicted in Figure \ref{fig:mccoderdemomachine}, to evaluate MCCoder's capabilities. The setup comprises a Windows PC running the MCCoder system, which interacts with various cloud-based LLMs and features a soft-motion system with both simulation and real-time engines, enabling code validation on the machine.
The demo setup also integrates precision linear motor stages with five axes and a six-axis collaborative robot (Cobot from Schneider). The electrical hardware includes EtherCAT servo drives for precise motion control and an I/O module featuring 16 digital inputs and 16 digital outputs, all of which facilitate comprehensive validation of the generated code.

\subsection{Base Models}

LLMs are advancing rapidly, consistently setting new state-of-the-art benchmarks. For this study, we selected the latest chat and reasoning LLMs as base models, as outlined in \Cref{tab:models}. GPT-4o, o3-mini, DeepSeek-V3, and DeepSeek-R1 were chosen for their proven excellence in code generation tasks. 
Reasoning models were included for their robust logical problem-solving and deep contextual understanding, essential for accurate and efficient automated code generation.

\begin{table}[t]
    \caption{Base Models for experiments}
    \label{tab:models}
    \centering
    \resizebox{\columnwidth}{!}{
        \begin{tabular}{cccc}
            \toprule
            Model    & Context Window & Max Output Tokens &  Params \\
            \midrule
            gpt-4o         & 200,000  & 16,384 & -   \\
            o3-mini    & 200,000 & 100,000 & -   \\
            \midrule
            DeepSeek-V3     & 64,000 & 8,192 & 671B   \\
            DeepSeek-R1     & 64,000 & 8,192 & 671B   \\
            \bottomrule
        \end{tabular}
    }
\end{table}

\subsection{Baseline}
Since motion control engineers nowadays frequently leverage chat-based models such as ChatGPT for code generation assistance, we designed our baseline to reflect this practical scenario. Specifically, we adopted an Advanced RAG approach, integrating hybrid retrieval and reranking to enhance contextual accuracy. This baseline serves as a reference for comparing the performance of MCCoder on the MCEVAL dataset.

\subsection{Experimental Results}


\begin{table*}[h]
    \caption{Experiments results of various models in MCEVAL datasets. For each model, the baseline Advanced RAG and the MCCoder system were compared. Metrics include FTPR (MatchEndPoints) and FTPR (DTW), with the latter being more stringent. The overall first-time pass rate was assessed and performance was evaluated at three difficulty levels. L1, L2, and L3. The best performance are \textbf{boldfaced} for Advanced RAG, and \underline{\textbf{underlined}} for MCCoder.}
    \label{tab:result}
    \small
    \centering
        \begin{tabular}{ccccccccccc}
            \toprule
            \multirow{2}{*}{Model} & \multirow{2}{*}{Method} & \multicolumn{4}{c}{FTPR(MatchEndPoints) $\uparrow$}  & & \multicolumn{4}{c}{FTPR(DTW) $\uparrow$} \\
            \cline{3-6} \cline{8-11}
            \rule{0pt}{1em} 
            & & OVERRALL & \footnotesize L1 & L2 & L3 & & \small OVERRALL & \footnotesize L1 & L2 & L3\\  
            \midrule
            
            \multirow{2}{*}{gpt-4o}  & Advanced RAG & 52.69  & \footnotesize 70.24 & \footnotesize 46.43 & \footnotesize 28.26 & & \small \textbf{63.98} & \footnotesize \textbf{84.52} & \footnotesize \textbf{55.36} & \footnotesize \textbf{36.96}  \\
             & \small MCCoder & \underline{\textbf{82.80}}  & \footnotesize 80.95 & \footnotesize \underline{\textbf{80.36}} & \footnotesize \underline{\textbf{89.13}} & & \underline{\textbf{69.35}} & \footnotesize 69.05 & \footnotesize \underline{\textbf{71.43}} & \footnotesize \underline{\textbf{67.39}}  \\
             \midrule
             
            \multirow{2}{*}{o3-mini}    & Advanced RAG & 55.38  & \footnotesize 70.24 & \footnotesize 55.36 & \footnotesize 28.26 & &  44.09 & \footnotesize 60.71 & \footnotesize 42.86 & \footnotesize 15.22  \\
            & \small MCCoder & 72.58  & \footnotesize 80.95 & \footnotesize 76.79 & \footnotesize 52.17 & & 60.75 & \footnotesize 71.43 & \footnotesize 67.86 & \footnotesize 32.61  \\
            
            \toprule
            
            \multirow{2}{*}{DeepSeek-V3}     & Advanced RAG & 65.05  & \footnotesize \textbf{88.10} & \footnotesize 57.14 & \footnotesize 32.61 & & 53.23 & \footnotesize 73.81 & \footnotesize 46.43 & \footnotesize 23.91  \\
               & \small MCCoder & 81.72  & \footnotesize 89.29 & \footnotesize 75.00 & \footnotesize 76.09 & & 65.59 & \footnotesize 72.62 & \footnotesize 64.29 & \footnotesize 54.35  \\
            \midrule
            \multirow{2}{*}{DeepSeek-R1}    & Advanced RAG & \textbf{66.67}  & \footnotesize 86.90 & \footnotesize \textbf{62.50} & \footnotesize \textbf{34.78} & & 56.45 & \footnotesize 76.19 & \footnotesize 51.79 & \footnotesize 26.09  \\
            & \small MCCoder & 79.03  & \footnotesize \underline{\textbf{90.48}} & \footnotesize 71.43 & \footnotesize 67.39 & & 66.13 & \footnotesize \underline{\textbf{78.57}} & \footnotesize 58.93 & \footnotesize 52.17  \\
                
            \bottomrule
        \end{tabular}

\end{table*}

We compared Advanced RAG and MCCoder using FTPR (MatchEndPoints) and FTPR (DTW) metrics. The overall pass rates and three difficulty levels were also evaluated for each model, as shown in \Cref{tab:result}. We will present results with three research questions.

\subsubsection{RQ1. Does MCCoder improve the code generation performance?}
Taking FTPR (MatchEndPoints) as an example, MCCoder achieved an overall FTPR improvement of 33.09\% compared to the baseline. Specifically, the improvements for Difficulty levels 1, 2, and 3 were 8.99\%, 39.33\%, and 131.77\%, respectively. This indicates that MCCoder significantly outperforms Advanced RAG, especially for more complex motion tasks. In other words, MCCoder demonstrates superior performance compared to the approach where control engineers manually provide sample codes to large models for code generation.

\subsubsection{RQ2. Which base model behaves the best with MCCoder?} 
MCCoder achieved the best overall performance when using gpt-4o as the base model, excelling in both matching end points and entire trajectory. The overall ranking of MCCoder's performance across different base models is: gpt-4o\textgreater{}DeepSeek-V3\textgreater{}DeepSeek-R1\textgreater{}o3-mini.
DeepSeek-V3 and R1 delivered results very close to gpt-4o, even surpassing it in some specific sections. However, when self-correction was required due to errors in generated answers, the Chain-of-Thought (CoT) revealed that the reasoning model tended to overthink in these specific motion control tasks. This led to an increase in both redundancy and error rates in the corrected outputs.

Meanwhile, o3-mini demonstrated the weakest overall performance, possibly due to its smaller model size, which may have limited its ability to handle the complexity of control tasks effectively.

\subsubsection{RQ3. Why is MCCoder's data verification critical for motion control?} 

In general-purpose Python programming, verification typically focuses on execution results via unit tests. In motion control, however, engineers must evaluate the entire execution process, including trajectory and endpoints, to ensure correctness and safety. Traceable data that clearly represent the entire process is essential.

\begin{figure}[t]
    \centering
    \includegraphics[width=1\columnwidth]{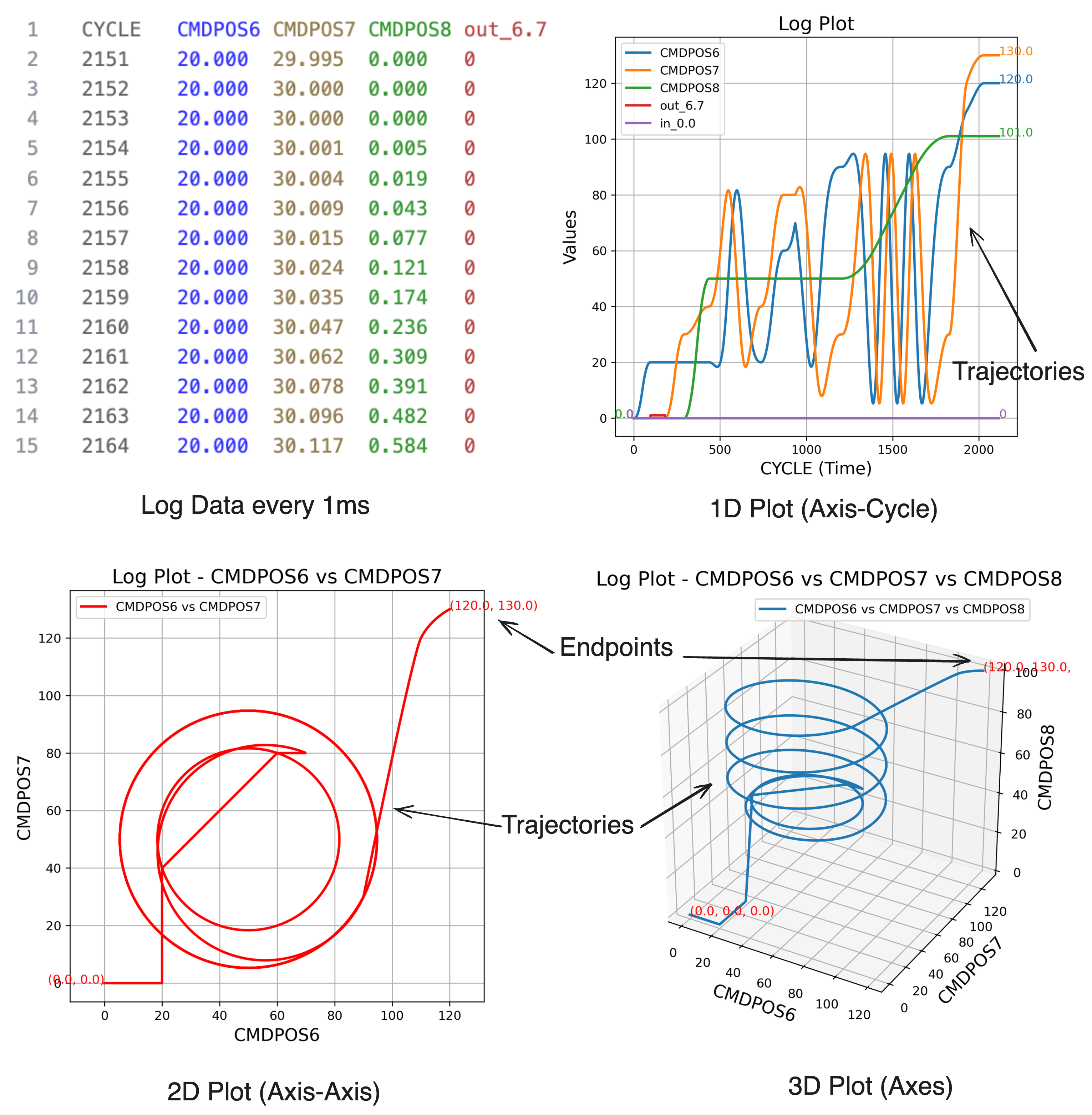}
    \vspace{-20pt}
    \caption{Example of logging data and plots from MCCoder system, providing data and visual virification for users.}
    \label{fig:MCEVAL116}
    \vspace{-10pt}
\end{figure}

MCCoder employs two verification approaches: instinct-driven verification through a real-time 3D simulator of the machine model and data-driven verification via logging files that capture the entire execution as time-series data, including both endpoints and trajectories. Additionally, 1D-3D trajectory plots of Axes and IOs provide a multimodal, traceable method to ensure the process aligns with expectations, as shown in \Cref{fig:MCEVAL116}.

\subsection{Error Analysis}
During the evaluation of the control codes, the soft-motion engine returns some errors. We analyze the errors in code generation with gpt-4o and their causes in detail.
\begin{enumerate}
    \item API errors (8.2\%): These errors indicate incorrect API function invoking. But the API error rate is relatively small, showing the effectiveness of the current retrieval method for relevant sample codes reference.
    \item Argument errors (76.3\%): The most prevalent error type involves correct API invocations with incorrect arguments, frequently stemming from hallucinations or disregard for prompt instructions. Self-correction can sometimes address and correct these errors. Explicit error messages like ``ProfileType has no attribute s curve. Did you mean: SCurve?" significantly enhance the likelihood of successful correction in subsequent attempts. In contrast, implicit error messages hinder accurate identification and correction.
    \item Syntax errors (15.5\%): These errors predominantly arise from hallucinations or the invention of nonexistent functions and libraries, resulting in syntactic inaccuracies.
\end{enumerate}

Self-correction is designed to address these returned errors efficiently. For general syntax errors, the LLM attempts direct self-repair. When errors involve APIs or arguments and originate from the Soft-Motion engine, the LLM further leverages error messages as references by searching related documentation to guide correction. These processes are typically fast, but to ensure time efficiency and smooth operation of the entire MCCoder system, the number of self-correction attempts is deliberately limited, for example, to three retries.

\section{DISCUSSION}


\subsection{Limitations and Future Work}

While MCCoder’s overall architecture and design can be applied to programming motion control tasks across various motion libraries using high-level languages, its current validation has been limited to the WMX3 library with Python. Future work will focus on expanding MCCoder’s applicability by collaborating with more motion control manufacturers and incorporating additional programming languages such as C++ and C\#. This will enhance MCCoder’s versatility and extend the applicability of MCEVAL’s evaluation data across a broader range of motion control systems.

The broader adoption of AI-assisted systems in industrial applications requires not only advancements in AI itself but also improvements in controller components. A key limitation is the need for more detailed error reporting and real-time interfaces, which are crucial for enhancing system efficiency and robustness.

A limitation of the current work is the lack of evaluation on more models, especially smaller-parameter ones, and the absence of systematic assessment of whether a well-designed workflow could enable small models to approach large-model performance. However, deploying MCCoder in real-world industrial field environments necessitates fine-tuned, edge-side small models tailored to proprietary motion libraries. These small models generate code faster than cloud-based large models and offer stronger data security, which is crucial on site. Fine-tuning ensures solid basic automation programming capabilities, while RAG allows dynamic incorporation of the latest code, templates, or user-defined patterns, achieving large-model-level programming advancements and further reducing hallucinations—making this a key research direction.

\section{CONCLUSION}

In this paper, we introduce MCCoder to leverage LLMs for generating motion control code. MCCoder addresses the complexities and safety-critical aspects of motion control programming by integrating prompt engineering, task decomposition, retrieval, code generation, soft-motion and verification through simulation and data logging. Validated using the MCEVAL dataset, MCCoder significantly improves code generation performance, especially for the most challenging tasks. These advancement provide valuable insights and guidelines for future deployment and research in industrial automation and control code generation. Future work will focus on enhancing MCCoder's versatility, improving error message clarity from the soft-motion controller, and exploring fine-tuning with edge-side small models to further enhance its capabilities and practicability.




\section*{ACKNOWLEDGMENT}

This work is funded by National Natural Science Foundation of China Grant No. 72371217, the Guangzhou Industrial Informatic and Intelligence Key Laboratory No. 2024A03J0628, the Nansha Key Area Science and Technology Project No. 2023ZD003, and Project No. 2021JC02X191.


\bibliographystyle{IEEEtran} 
\bibliography{root}

\end{document}